\def\year{2018}\relax
\documentclass[letterpaper]{article} 
\usepackage{aaai18}  
\usepackage{times}  
\usepackage{helvet}  
\usepackage{courier}  
\usepackage{url}  
\usepackage{graphicx}  

\usepackage{amsfonts}
\usepackage{algorithm}
\usepackage{algorithmic}
\usepackage{wrapfig}
\usepackage{amsmath}
\usepackage{epstopdf}
\input xy
\xyoption{all}
\usepackage{subfig}
\usepackage{tabularx}
\usepackage{mathrsfs} 
\usepackage{times}
\usepackage{graphicx} 
\usepackage{algorithm}
\usepackage{algorithmic}
\usepackage{comment}
\usepackage{amsthm}
\usepackage{mathtools}
\usepackage{grffile}
\usepackage{etex}

\usepackage[colorinlistoftodos,prependcaption]{todonotes}
\newtheorem{Theorem}{Theorem}
\newtheorem{Lemma}{Lemma}
\newtheorem{proposition}{Proposition}

\newtheorem{Definition}{Definition}

\newtheorem{assumption}{Assumption}

\begin{document}
%
\title{Learning Vector Autoregressive Models with Latent Processes}
\author{Saber Salehkaleybar$^*$, Jalal Etesami$^{*\dagger}$, Negar Kiyavash$^{\dagger\ddagger}$, Kun Zhang$^{\Diamond}$\\
$^*$Coordinated Science Laboratory, University of Illinois at Urbana-Champaign, Urbana, USA.\\
$^{\dagger}$Department of ISE, University of Illinois at Urbana-Champaign
Urbana, USA.\\
$^{\ddagger}$Department of ECE, University of Illinois at Urbana-Champaign, Urbana, USA.\\
$^{\Diamond}$Department of Philosophy, Carnegie Mellon University, Pittsburgh, USA.\\
\texttt{\{sabersk,etesami2,kiyavash\}@illinois.edu,kunz1@cmu.edu}\\
}
\maketitle
\begin{abstract}

We study the problem of learning the support of transition matrix between random processes in a Vector Autoregressive (VAR) model from samples when a subset of the processes are latent. It is well known that ignoring the effect of the latent processes may lead to very different estimates of the influences among observed processes, and we are concerned with identifying the influences among the observed processes, those between the latent ones, and those from the latent to the observed ones. We show that the support of transition matrix among the observed processes and lengths of all latent paths between any two observed processes can be identified successfully under some conditions on the VAR model. From the lengths of latent paths, we reconstruct the latent subgraph (representing the influences among the latent processes) with a minimum number of variables uniquely if its topology is a directed tree. Furthermore, we propose an algorithm that finds all possible minimal latent graphs under some conditions on the lengths of latent paths. Our results apply to both non-Gaussian and Gaussian cases, and experimental results on various synthetic and real-world datasets validate our theoretical results.

\end{abstract}

\section{Introduction}
Identifying causal influences among time series is a problem of interest in many fields. In macroeconomics, for instance, researchers seek to understand what factors contribute to economic fluctuations and how they interact with each other \cite{Lutkepohl04}. In neuroscience, many researchers focus on learning the interactions between different regions of brain by analyzing neural spike trains \cite{roebroeck2005mapping,besserve2010causal,di12}.

Granger causality \cite{Granger}, transfer entropy \cite{schreiber2000measuring}, and directed information \cite{massey1990causality,Marko} are some of the most commonly used measures in the literature to calculate time-delayed dependence structures in time series. 
Measuring the reduction of uncertainty in one variable after observing another variable is the key concept behind such measures. Under certain assumptions, these measures may represent causal relations among the variables \cite{pearl2009causality,spirtes2000causation}. In \cite{eichler2012causal}, an overview of various definitions of causation is given for time series.

In this work, we study the causal identification problem in VAR models when only a subset of times series is observed.  
More precisely, we assume that the available measurements are a set of random processes $\vec{X}(t)\in\mathbb{R}^{n}$ which, together with another set of latent random processes $\vec{Z}(t)\in\mathbb{R}^{m}$, where $m\leq n$ form a first order VAR model as follows:

\begin{equation}\label{eq:model}
\begin{bmatrix}
\vec{X}(t+1)\\\vec{Z}(t+1)
\end{bmatrix}
=\begin{bmatrix}
A_{11}& A_{12}\\
A_{21} & A_{22}
\end{bmatrix}
\begin{bmatrix}
\vec{X}(t)\\
\vec{Z}(t)
\end{bmatrix}
+
\begin{bmatrix}
\vec{\omega}_X(t+1)\\
\vec{\omega}_Z(t+1)
\end{bmatrix}.
\end{equation}

Here we assume that observed data were measured at the right causal frequency of the VAR process; otherwise one may need to consider the effect of the sampling procedure such as subsampling or temporal aggregation~\cite{Danks14,gong2015discovering,gong2017UAI}. Under certain assumptions (e.g., causal sufficiency), the support of the transition matrix corresponds to the causal structure between these processes \cite{Granger,spirtes2000causation,pearl2009causality}. 
If we ignore the influence of latent processes and just regress $\vec{X}(t+1)$ on $\vec{X}(t)$, we may get a wrong estimate of the transition matrix between observed processes (see the example in \cite{Geiger}). Hence, it is crucial to consider the presence of latent processes and their influences on the observed processes.

%

{\bf Contributions:} The contributions of this paper are as follows: we propose a learning approach that recovers the \textit{observed sub-network} (support of $A_{11}$) by regressing the observed vector $\vec{X}(t+1)$ on a set of its past observations (not just $\vec{X}(t)$) as long as the graph representation of \textit{latent sub-network} (support of $A_{22}$) is a directed acyclic graph (DAG).
We also derive a set of sufficient conditions under which we can uniquely recover the influences from latent to observed processes, (support of $A_{12}$) and also the influences among the latent variables, (support of $A_{22}$). 
Additionally, we propose a sufficient condition under which the support of the complete transition matrix can be recovered uniquely. 

More specifically, we show that under an assumption on the observed to latent noise power ratio, if neither of the sub-matrices $A_{12}$ and $A_{21}$ are zero,
it is possible to determine the length of all directed \textit{latent paths}\footnote{A directed path is a latent path if it connects two observed variables and all the intermediate variables on that path are latent.}. 
We refer to this information as \textit{linear measurements}\footnote{This is because it can be inferred from the observational data using linear regression.}.
This information reveals important properties of the causal structure among the latent and observed processes, i.e., support of $[0, A_{12}; A_{21}, A_{22}]$.
 We call this sub-network of a VAR model \textit{unobserved network}. 
 We show that in the case that the unobserved network is a directed tree and each latent variable has at least two parents and two children, a straightforward application of \cite{Hakimi} can recover the unobserved network uniquely.
Furthermore, we propose Algorithm \ref{AlgDTR} that recovers the support of $A_{22}$ and $A_{12}$ given the linear measurements when only the latent sub-network is a directed tree plus some extra structural assumptions (see Assumption \ref{ass1}).
Lastly, we study the causal structures of VAR models in a more general case in which there exists at most one directed latent path of length $k\geq2$ between any two observed processes (see Assumption \ref{ass2}).
For such VAR models, we propose Algorithm \ref{AlgNM} that can recover all possible unobserved networks with minimum number of latent processes. Our results apply to both non-Gaussian and Gaussian cases, and experimental results on various synthetic and real-world datasets validate our theoretical results. All proofs can be found in supplemental material.

{\bf Related works:} 
The problem of recovering latent causal structure for time series has been studied in the literature. Assuming that connections between observed variables are sparse and each latent variable interacts with many observed variables, it has been shown 
 that the transition matrix between observed variables can be identified in a VAR model \cite{Jalali}. However, their approach focuses on learning only the observed sub-network.
\cite{boyen1999discovering} applied a method based on expectation maximization (EM) to infer properties of partially observed Markov processes, without providing theoretical analysis for identifiability. 
\cite{Geiger} showed that if the exogenous noises are independent non-Gaussian and additional so-called genericity assumptions hold, then the sub-networks $A_{11}$ and a part of $A_{12}$ are uniquely identifiable. However, these assumptions may not hold true in a real-world dataset even with three variables \cite{Geiger}.
They also presented a result in which they allowed Gaussian noises in their VAR model and obtained a set of conditions under which they can recover  up to ${2n \choose n}$ candidate matrices for $A_{11}$.
Their learning approach is also based on EM and approximately maximizes the likelihood of a parametric VAR model with a mixture of Gaussians as noise distribution.
Recently, \cite{Etesami} studied a network of processes (not necessary a VAR model) whose underlying structure is a polytree and introduced an algorithm that can learn the entire casual structure (observed and unobserved networks) using a particular discrepancy measure.

Compared to related works, we improve the state of the art for latent recovery by showing the identifiability of a much larger class of structures. Unlike \cite{Geiger}, we do not assume the non-Gaussian distribution of the exogenous noises or those genericity assumptions. Moreover, our results do not rely on the assumption that connections between observed variables are sparse or each latent variables interacts with many observed variables as in \cite{Jalali}. Furthermore, these works \cite{Geiger,Jalali} can uniquely identify at most a part of transition matrix ($A_{11}$ or a part of $A_{12}$). 
\vspace{-3mm}
\section{Problem Definition}
\vspace{-1mm}

In this part, we review some basic definitions and our notation. 
Throughout this paper, we use an arrow over the letters to denote vectors.
We assume that the time series are stationary and denote the autocorrelation of $\vec{X}$ by $\gamma_X(k):=\mathbb{E}[\vec{X}(t)\vec{X}(t-k)^T]$. 
We denote the support of a matrix $A$ by $Supp(A)$ and use $Supp(A)\subseteq Supp(B)$ to indicate $[A]_{ij}=0$ whenever $[B]_{ij}=0$.
We also denote the Fourier transform of $g$ by $\mathcal{F}(g)$ and it is given by \begin{small}$\sum_{h=-\infty}^{\infty} g(h) e^{-h\Omega j}$\end{small}.

In a directed graph $G=(V,\overrightarrow{E})$ with the node set $V$ and the edge set $\overrightarrow{E}$, we denote the set of parents of a node $v$ by $\mathcal{P}_v:=\{u: (u,v)\in\overrightarrow{E}\}$ and the set of its children by $\mathcal{C}_v:=\{u: (v,u)\in\overrightarrow{E}\}$. 
The skeleton of a directed graph $G$ is the undirected graph obtained by removing all the directions in $G$.

\vspace{-3mm}
\label{PD}
\subsection{System Model}

Consider the VAR model in (\ref{eq:model}). Let $\vec{\omega}_X(t)\in\mathbb{R}^n$ and $\vec{\omega}_Z(t)\in\mathbb{R}^m$ be i.i.d random vectors with mean zero. For simplicity, we denote the matrix $[A_{11},A_{12};A_{21},A_{22}]$ by $A$. 
Our goal is to recover $Supp(A)$ from observational data, i.e., $\{\vec{X}(t)\}$. Rewrite \ref{eq:model} as follows
 \vspace{-.1cm}
\begin{small}
\begin{align}\label{eqmain2}\notag
\vec{X}(t+1)=&\displaystyle\sum_{k=0}^t A^*_k\vec{X}(t-k) + A_{12}A_{22}^t \vec{Z}(0)+\\&\displaystyle\sum_{k=0}^{t-1} \tilde{A}_k\vec{\omega}_Z(t-k) +\vec{\omega}_X(t+1), 
\end{align}
\end{small}where $A_0^*:=A_{11}$, $A_k^*:=A_{12}A_{22}^{k-1}A_{21}$ for $k\geq 1$, and $\tilde{A}_k:= A_{12}A_{22}^k$. 

\begin{assumption}
We assume that the $A_{22}$ is acyclic, i.e., $\exists\ 0<l\leq m$, such that $A_{22}^l=0$.
\label{ass0}
\end{assumption}

Based on the above assumption, for $t\geq l$, Equation \eqref{eqmain2} becomes\footnote{Note that the limits of summations in (\ref{eqfirst}) are changed.}
 \begin{small}
\begin{equation}\label{eqfirst}
\vec{X}(t\!+1)=\!\sum_{k=0}^l A^*_k\vec{X}(t\!-k)\! +\! \displaystyle\sum_{k=0}^{l-1} \tilde{A}_k\vec{\omega}_Z(t\!-k)+\vec{\omega}_X(t\!+1).
\end{equation}
\end{small}We are interested in recovering the set $\{Supp(A_k^*)\}_{k=0}^l$ because it captures important information about the structure of the VAR model. 
Specifically, $Supp(A^*_{0})=Supp(A_{11})$; so it represents the direct causal influences between the observed variables and $Supp(A^*_k)$ for $k\geq1$ determines whether at least one directed path of length $k+1$ exists between any two observed nodes which goes through the latent sub-network.\footnote{Herein, we exclude degenerate cases where there is a direct path from an observed node to another one with length $k$ but the corresponding entry in matrix $Supp(A^*_k)$ is zero. In fact, such special cases can be resolved by small perturbation of nonzero entries in matrix $A$. In the causal discovery literature, this assumption is known as faithfulness~\cite{spirtes2000causation}.} 
We will make use of this information in our recovery algorithm.
We call the set of matrices $\{Supp(A_k^*)\}_{k\geq0}$, \textit{linear measurements}.
In Section 4, we present a set of sufficient conditions under which given the linear measurements, we can recover the entire or most parts of the unobserved network uniquely. 
\ref{sec:learn}

\begin{figure}
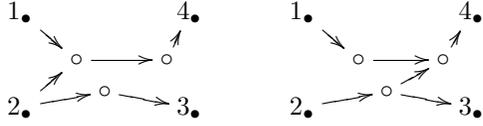

	\vspace{-.3cm}
	\hspace{.3cm}
\xygraph{ !{<0cm,0cm>;<1.5cm,0cm>:<0cm,0.7cm>::} 
!{(-2,1.8) }*+{1_\bullet}="1"   
  !{(-2,0) }*+{2_\bullet}="2"	
	!{(-0.5,0) }*+{3_\bullet}="3"   	 		
!{(-0.5,1.8) }*+{4_\bullet}="4" 
!{(-1.5,0.9) }*+{\circ}="l1"  
!{(-0.7,0.9) }*+{\circ}="l2" 
!{(-1.25,0.3) }*+{\circ}="l3" 
!{(0.5,1.8) }*+{1_\bullet}="1p"   
  !{(0.5,0) }*+{2_\bullet}="2p"	
	!{(2,0) }*+{3_\bullet}="3p"   	 		
!{(2,1.8) }*+{4_\bullet}="4p" 
!{(1,0.9) }*+{\circ}="l1p"  
!{(1.75,0.9) }*+{\circ}="l2p" 
!{(1.25,0.3) }*+{\circ}="l3p" 
"1":"l1"  "l1":"l2"  "l2":"4" "2":"l3"  "l3":"3" "2":"l1"
"1p":"l1p" "l1p":"l2p" "l2p":"4p" "2p":"l3p" "l3p":"l2p" "l3p":"3p"
} 
\caption{Two unobserved networks with the same linear measurements. White circles denote latent nodes.}\label{Fig1}
	\vspace{-.6cm}
\end{figure}

Note that in general, the linear measurements cannot uniquely specify the unobserved network. 
For example, Figure \ref{Fig1} illustrates two different unobserved networks that both share the same set of linear measurements, $A^*_k=0$ for $k>2$ and the only nonzero entries of $A^*_1$ and $A^*_2$ are $\{(3,2)\}$ and $\{(4,1),(4,2)\}$, respectively.
\vspace{-3mm}
\section{Identifiability of the Linear Measurements}\label{sec:lin}
As we need the linear measurements for our structure learning, in this section, we study a sufficient condition under which we can recover the linear measurements from the observed processes $\{\vec{X}(t)\}$.
To do so, we start off by rewriting Equation (\ref{eqfirst}) as follows,
 \begin{small}
\begin{equation}\label{eqmatrixform}
\vec{X}(t+1)=\mathcal{A}\vec{\mathcal{X}}_{t-l:t}+\sum_{k=0}^{l-1} \tilde{A}_k\vec{\omega}_Z(t-k)+\vec{\omega}_X(t+1),
\end{equation}
 \end{small}where \begin{small}$\mathcal{A}:= [A_0^*,...,A_l^*]$\end{small}, and \begin{small}$\vec{\mathcal{X}}_{t-l:t}:=[\vec{X}(t);\cdots;\vec{X}(t-l)].$\end{small}
By projecting $\tilde{A}_k\vec{\omega}_Z(t-k)$ onto the vector space spanned by the observed processes, i.e., $\{\vec{X}(t),...,\vec{X}(t-l)\}$, we obtain
\begin{small}
\begin{equation}\label{eq11}
\tilde{A}_k\vec{\omega}_Z(t-k)\!=\!\sum_{r=0}^{l} C_r^s \vec{X}(t-r) + \vec{N}_Z(t-k),  \ 0\leq\! k\!\leq l\!-1,
\end{equation}
\end{small}where $\{\vec{N}_Z(t-k)\}$ denote the residual terms and $\{C_r^s\}$ are the corresponding coefficient matrices.
Substituting (\ref{eq11}) into (\ref{eqmatrixform}) implies
\begin{small}
\begin{equation}\label{matrixform}
\vec{X}(t+1)=\mathcal{B}\vec{\mathcal{X}}_{t-l:t}+\vec{\theta}(t+1),
\end{equation}
\end{small}where $\mathcal{B}:=[B^*_0,...,B^*_{l}]$, 
\begin{small}
$
B^*_k:=A_k^*+\sum_{s=0}^{l-1} C_k^s,$ and  $\vec{\theta}(t+1):=\vec{\omega}_X(t\!+\!1)\!+\!\sum_{k=0}^{l-1} \vec{N}_Z(t\!-\! k).
$
\end{small}
Note that by this representation, $\vec{\theta}(t+1)$ is orthogonal to $\vec{\mathcal{X}}_{t-l:t}$.
Hence, Equation (\ref{matrixform}) shows that the minimum mean square error (MMSE) estimator 
can learn the coeffiecient matrix $\mathcal{B}$ given the observed processes. More precisely, let $\Gamma_X(l):= \mathbb{E}\{\vec{\mathcal{X}}_{t-l:t} \vec{\mathcal{X}}_{t-l:t}^T\}$, then we have 
\begin{equation}\label{eq:es1}
\mathcal{B}=[\gamma_X(1),..,\gamma_X(l+1)]\times\Gamma_{X}(l)^{-1}.
\end{equation}
\begin{proposition}\label{prop1}
Under Assumption \ref{ass0}, for the stationary VAR model in (\ref{eq:model}), we have
\begin{equation}\notag
||B^*_k-A_k^*||_1\leq \sqrt{n(l\!-\!k\!-\!1) M/L}||A_{12}||_2||A_{22}||_2^{k+1},
\end{equation}
where 
$M:=\lambda_{max}(\Gamma_{\omega_Z}(0))$ and $L:=\lambda_{min}\left(\Gamma_{X}(0)\right)$. 
\end{proposition} 
This result implies that we can asymptotically recover the support of  $\{A_k^*\}_{k=0}^l$ as long as the absolute values of non-zero entries of $A_k^*$ are bounded away from zero by $2\sqrt{n(l\!-\!k\!-\!1)\frac{M}{L}}||A_{12}||_2||A_{22}||_2^{k+1}$. Please note that $A_{11}=A_{0}^*=B_0^*$ if $||A_{12}||_2\!=\!0$.
In Appendix (the second section), we explained how these bounds can be estimated from observational data.

\begin{proposition}\label{propSNR}
Let $\Sigma_X=\sigma^2_X I_{n\times n}$ and $\Sigma_Z=\sigma_Z^2 I_{m\times m}$ be the autocovariance matrices of $\vec{\omega}_X(t)$ and $\vec{\omega}_Z(t)$, respectively. Then, the ratio $M/L$ strictly increases by decreasing $\sigma_X^2/\sigma_Z^2$.
\end{proposition}
Proposition \ref{propSNR} implies that when the $\sigma_X^2/\sigma_Z^2$ increases, $M/L$ will decrease, and based on the bound in Proposition \ref{prop1}, the estimation error will decrease (it goes to zero as $\sigma_X^2/\sigma_Z^2$ tends to infinity). This shows that recovering the linear measurements is much easier in high $\sigma_X^2/\sigma_Z^2$ regime as illustrated in Figure \ref{fig6b}. Note that Proposition \ref{prop1} stresses a suffiecient condition for recovering the linear measurements. As shown in Figure \ref{fig6b}, in practice, the actual estimation error is much smaller than the bound in Proposition \ref{prop1}. In the next section, we will make use of $\{Supp(A_k^*)\}_{k>0}$ to recover the unobserved network. We assume that the correct linear measurements can be obtained from matrix $\mathcal{B}$.

In order to estimate the support of matrix $\mathcal{B}$ from a finite number of samples drawn from the observed processes, say $\{\vec{X}(t)\}_{t=1}^T$, first we obtain the lag length $l$ in \eqref{matrixform} by AIC or FPE criterion (see Chapter 4 in \cite{lutkepohl2005new}). Afterwards, we can estimate the coefficient matrix $\mathcal{B}$, using an empirical estimator for $\Gamma_{X}(l)$, $\{\gamma_{X}(h)\}_{h=1}^{l+1}$, and then applying (\ref{eq:es1}). Denote the result of this estimation by $\mathcal{B}_T$. It can be shown that \cite{lutkepohl2005new}, 
\begin{small}
$
\sqrt{T}\text{vec}(\mathcal{B}_T-\mathcal{B})\xrightarrow[T\rightarrow \infty]{d} \mathcal{N}(0,  \Gamma^{-1}_{X}(l)\otimes\Sigma),
$
\end{small}
where $\xrightarrow{d}$ denotes convergence in distribution, and $\Sigma$ is the autocovariance matrix of $\vec{\theta}(t)$. $\text{vec}(.)$ transforms a matrix to a vector by stacking its columns and $\otimes$ is the Kronecker product. 
Having the estimates of $\Gamma_{X}(l)$ and $\Sigma$, we can test whether the entries of matrix $\mathcal{B}$ are greater than the bounds in Proposition \ref{prop1}  (see Chapter 3 in \cite{lutkepohl2005new}).


\vspace{-3mm}
\section{Learning the Unobserved Network}\label{sec:learn}
Recall that we refer to \begin{small}$Supp([0, A_{12}; A_{21}, A_{22}])$\end{small} as the unobserved network and \begin{small}$Supp(A_{22})$\end{small} as the latent sub-network. 
We present three algorithms that take the linear measurements \begin{small}$\{Supp(A^*_k)\}_{k\geq 0}$\end{small} as their input.
The first algorithm recovers the entire unobserved network uniquely as long as it is a directed tree and  each latent node has at least two parents and two children.
The output of the second algorithm is \begin{small}$Supp([0, A_{12}; \widehat{A}_{21}, A_{22}])$\end{small}, where \begin{small}$Supp(A_{21})\subseteq Supp(\widehat{A}_{21})$\end{small}. 
This is guaranteed whenever the latent sub-network is a directed tree and some extra conditions are satisfied on how the latent and observed nodes are connected.
The third algorithm finds the set of all possible networks with minimum number of latent nodes that are consistent with the measurements.
This algorithm is able to do so when there exists at most one directed latent path of any arbitrarily length between two observed nodes. A directed path is latent if all the intermediate variables on that path are latent.

\subsection{Unobserved Network is a Directed Tree}
Authors in \cite{Hakimi} introduced a necessary and sufficient condition for recovering a weighted directed tree uniquely from a valid distance matrix $D$ defined on the observed nodes,\footnote{The skeleton of the recovered tree is the same as the original one but not necessary the weights.}  and also proposed a recovery algorithm. 
The condition is as follows: every latent node must have at least two parents and two children. 
A matrix $D$, in \cite{Hakimi}, is a valid distance matrix, when $[D]_{ij}$ equals the sum of all the weights of those edges that belong to the directed path from $i$ to $j$, and $[D]_{ij}=0$, if there is no directed path.

The algorithm in \cite{Hakimi} has two phases. In the first phase, it creates a directed graph among the observed nodes with the adjacency matrix $Supp(D)$. 
In the second phase, it recursively finds and removes the circuits by introducing latent nodes for each circuit.\footnote{In a directed graph, a circuit is a cycle after removing all the directions.}
For more details, see \cite{Hakimi}.

%

In order to adopt \cite{Hakimi}'s algorithm for learning the unobserved network, we introduce a valid distance matrix using our linear measurements as follows, $D_{ij}=k+1$ if  $[Supp(A_k^*)]_{ji}\neq 0$ and 0, otherwise.
Recall that $[Supp(A_k^*)]_{ji}$ indicates whether there exists a directed latent path from $i$ to $j$ of length $k+1$ in the unobserved network. From theorem 8 in \cite{Hakimi}, it is easy to show that the unobserved network can be recovered uniquely from above distance matrix if its topology is a directed tree and every latent node has at least two parents and two children.
\vspace{-3mm}
\subsection{Latent Sub-network Is a Directed Tree}\label{sec:ff}
\begin{Definition}
We denote the subset of observed nodes that are parents of a latent node $h$ by $\mathcal{P}^O_h$ and denote the subset of observed nodes for which $h$ is a parent, by $\mathcal{C}^O_h$. We further denote the set of all leaves in the latent sub-network by $\mathcal{L}$.
\end{Definition}

We consider learning an unobserved network $G$ that satisfies the following assumptions.

\begin{assumption}\label{ass1}
Assume that the latent sub-network of $G$ is a directed tree. Furthermore, for any latent node $h$ in $G$, (i) $\mathcal{P}^O_h\not\subseteq\cup_{h\neq j}\mathcal{P}^O_j$ and, (ii) if $h$ is a leaf of the latent sub-network, then $\mathcal{C}^O_h\not\subseteq\cup_{ i\in\mathcal{L},i\neq h}\mathcal{C}^O_i$.
\end{assumption}

This assumption states that the latent sub-network of $G$ must be a directed tree such that each latent node in $G$ has at least one unique parent in the set of observed nodes. That is, a parent who is not shared with any other latent node. Furthermore, each latent leaf has at least one unique child among the observed nodes.
For instance, when $Supp(A_{22})$ represents a directed tree and both $Supp(A_{12})$ and $Supp(A_{21})$ contain identity matrices, Assumption \ref{ass1} holds. As we will see later in Experimental Results (Figure \ref{figPsat}), a large portion of randomly generated graphs satisfy Assumption \ref{ass1}.

\begin{figure*}
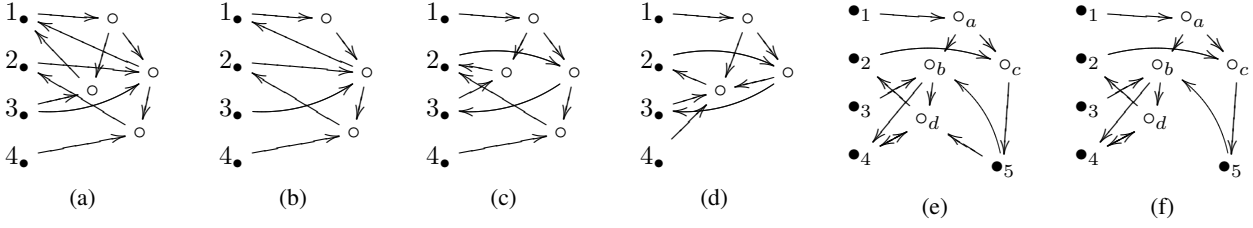

\subfloat[]{\label{v1}\xygraph{ !{<0cm,0cm>;<.9cm,0cm>:<0cm,.8cm>::} 
!{(.5,1.7) }*+{\circ}="1"   
  !{(.2,.5) }*+{\circ}="3"	 !{(1.1,.8) }*+{\circ}="23"  
	!{(.9,-.20) }*+{\circ}="4"    	 	   	 	
!{(-.9,1.8) }*+{1_\bullet}="x1" !{(-.9,1.) }*+{2_\bullet}="x2"  !{(-.9,.2) }*+{3_\bullet}="x3" !{(-.9,-.6) }*+{4_\bullet}="x4" 
"x1":"1"  "x3":"3"  "x2":"23" "x3":@/_/"23" "x4":"4"
"1":"3" "1":"23" "23":"4" "3":"x1"  "23":"x1" "4":"x2"
} }
\hspace{0.4cm}
\subfloat[]{\label{v2}
\xygraph{ !{<0cm,0cm>;<.9cm,0cm>:<0cm,.8cm>::} 
!{(.5,1.7) }*+{\circ}="1"   
 !{(1.1,.8) }*+{\circ}="23"  
	!{(.9,-.20) }*+{\circ}="4"    	 	   	 	
!{(-.9,1.8) }*+{1_\bullet}="x1" !{(-.9,1.) }*+{2_\bullet}="x2"  !{(-.9,.2) }*+{3_\bullet}="x3" !{(-.9,-.6) }*+{4_\bullet}="x4" 
"x1":"1"   "x2":"23" "x3":@/_/"23" "x4":"4"
 "1":"23" "23":"4"   "23":"x1" "4":"x2"
 } }
\hspace{0.4cm}
\subfloat[]{\label{w1}\xygraph{ !{<0cm,0cm>;<.9cm,0cm>:<0cm,.8cm>::} 
!{(.5,1.7) }*+{\circ}="1"   
  !{(.1,.8) }*+{\circ}="3"	 !{(1.1,.8) }*+{\circ}="2"  
	!{(.9,-.20) }*+{\circ}="4"    	 	   	 	
!{(-.9,1.8) }*+{1_\bullet}="x1" !{(-.9,1.) }*+{2_\bullet}="x2"  !{(-.9,.2) }*+{3_\bullet}="x3" !{(-.9,-.6) }*+{4_\bullet}="x4" 
"x1":"1"  "x3":"3"  "x2":@/^/"2"  "x4":"4"
"1":"2" "1":"3" "2":"4" "2":@/^/"x3" "3":"x2"  "4":"x2"
} }
\hspace{0.4cm}
\subfloat[]{\label{w2}
\xygraph{ !{<0cm,0cm>;<.9cm,0cm>:<0cm,.8cm>::} 
!{(.5,1.7) }*+{\circ}="1"   
  !{(.1,.5) }*+{\circ}="3"	 !{(1.1,.8) }*+{\circ}="2"  
!{(-.9,1.8) }*+{1_\bullet}="x1" !{(-.9,1.) }*+{2_\bullet}="x2"  !{(-.9,.2) }*+{3_\bullet}="x3" !{(-.9,-.6) }*+{4_\bullet}="x4" 
"x1":"1"  "x3":"3"  "x2":@/^/"2"  "x4":"3"
"1":"2" "1":"3" "2":"3" "2":@/^/"x3" "3":"x2" 
 } }
 \hspace{0.4cm}
 \subfloat[]{\label{z1}\xygraph{ !{<0cm,0cm>;<1cm,0cm>:<0cm,.8cm>::} 
!{(.5,1.7) }*+{\circ_a}="1"   
  !{(.1,.9) }*+{\circ_b}="3"	 !{(1.1,.9) }*+{\circ_c}="2"  
	!{(0,-.0) }*+{\circ_d}="4"   	 		
!{(-.9,1.8) }*+{\bullet_1}="x1" !{(-.9,1.) }*+{\bullet_2}="x2"  !{(-.9,.2) }*+{\bullet_3}="x3" !{(-.9,-.6) }*+{\bullet_4}="x4" 
 !{(1,-.8) }*+{\bullet_5}="x6"
"x1":"1"  "x3":"3"    "x4":"4"    "x2":@/^/"2"
"1":"3" "1":"2"  "3":"4" "3":"x4"
"4":"x2" "4":"x4"  "2":"x6" "x6":@/_/"3" "x6":"4"
} }
\hspace{0.4cm}
\subfloat[]{\label{z2}\xygraph{ !{<0cm,0cm>;<1cm,0cm>:<0cm,.8cm>::} 
!{(.5,1.7) }*+{\circ_a}="1"   
  !{(.1,.9) }*+{\circ_b}="3"	 !{(1.1,.9) }*+{\circ_c}="2"  
	!{(0,-.0) }*+{\circ_d}="4"   	 		
!{(-.9,1.8) }*+{\bullet_1}="x1" !{(-.9,1.) }*+{\bullet_2}="x2"  !{(-.9,.2) }*+{\bullet_3}="x3" !{(-.9,-.6) }*+{\bullet_4}="x4" 
 !{(1,-.8) }*+{\bullet_5}="x6"
"x1":"1"  "x3":"3"    "x4":"4"    "x2":@/^/"2"
"1":"3" "1":"2"  "3":"4" "3":"x4"
"4":"x2" "4":"x4"  "2":"x6" "x6":@/_/"3" 
} }
\caption{Latent nodes are indicated by white circles. Graph (a) satisfies (ii) but not (i) and it can be reduced to (b). Graph (c) satisfies (i) but not (ii) and it can be reduced to (d). (e) and (f) satisfy Assumption \ref{ass1} and have the same induced linear measurements but \small{$Supp(\! A_{21}\!)_{(f)}\!\subset\! Supp(\!A_{21}\!)_{(e)}$}.}\label{uu}
\vspace{-3mm}
\end{figure*}

\begin{figure*}
\def\tabularxcolumn#1{m{#1}}
\begin{tabularx}{\linewidth}{@{}cXX@{}}
\begin{tabular}{ccc}
\subfloat[The average normalized error versus number of observed nodes.]{\includegraphics[width=2in,height=1.5in]{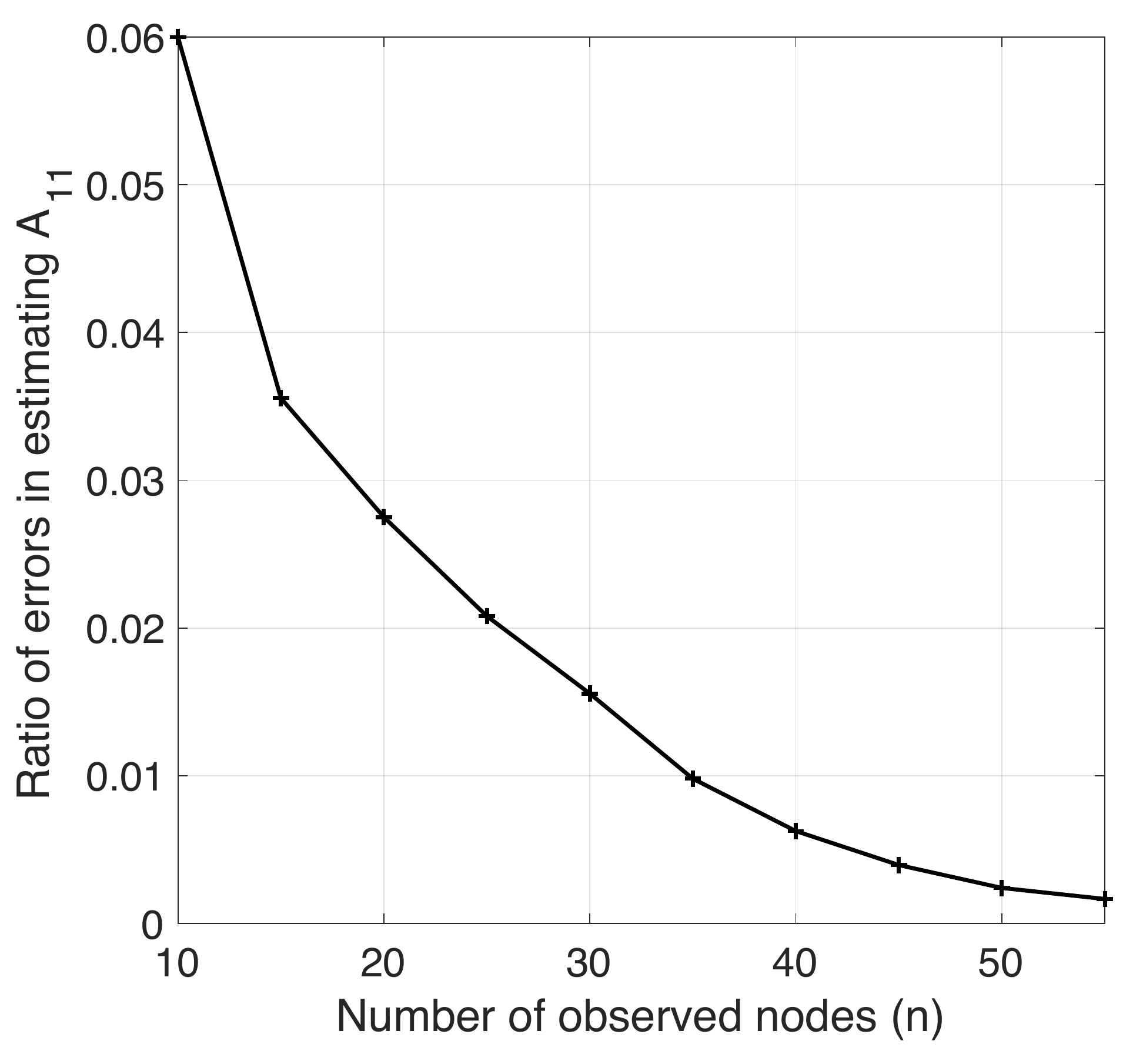}\label{fig6a}}
&
\subfloat[The average of estimation error versus OLNR.]{\hspace{0.4cm}\includegraphics[width=2in,height=1.5in]{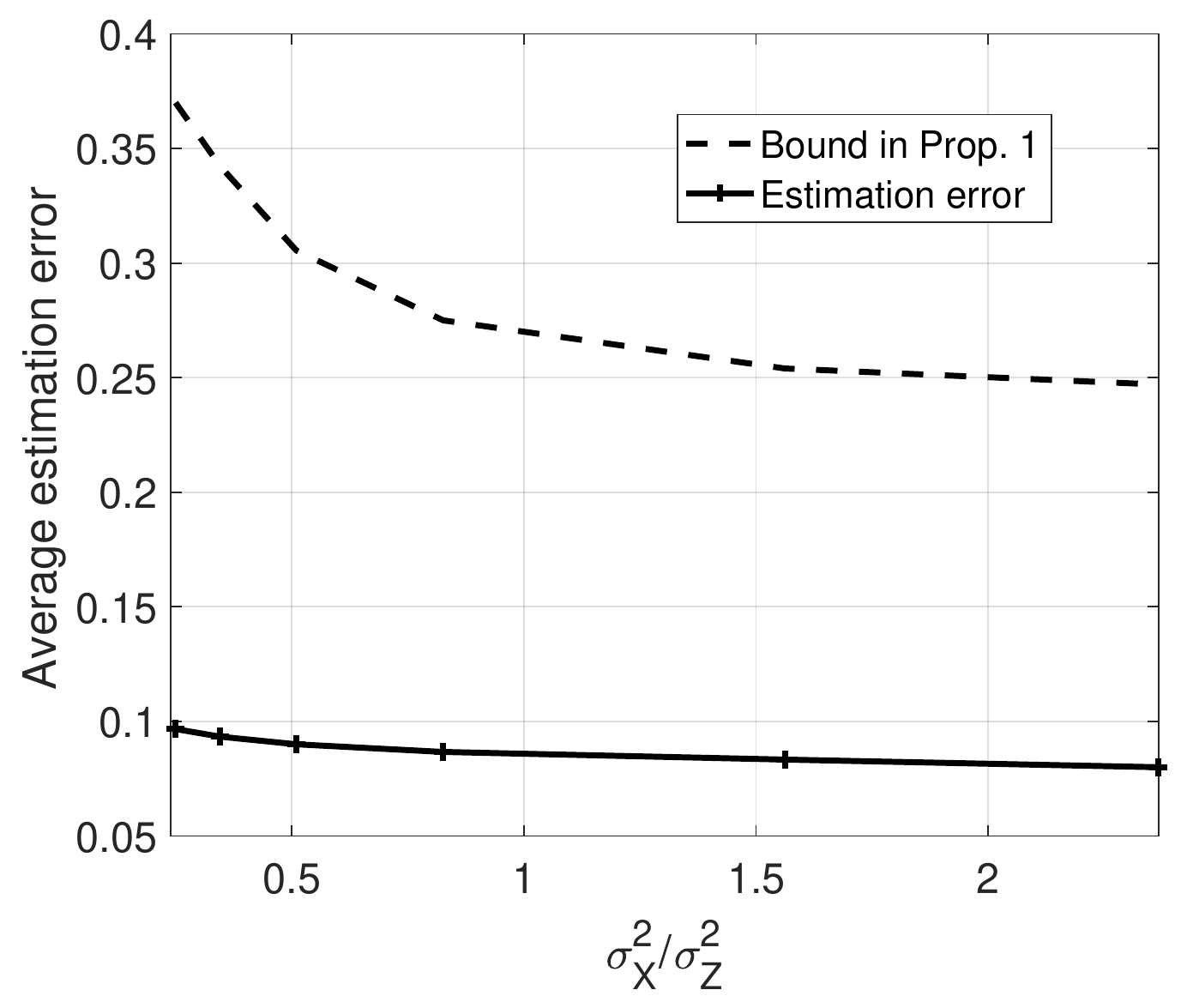}\label{fig6b}}
&
\subfloat[The probability $P_{sat.}$ versus the parameter $p$.]{\hspace{0.4cm}\includegraphics[width=1.8in,height=1.5in]{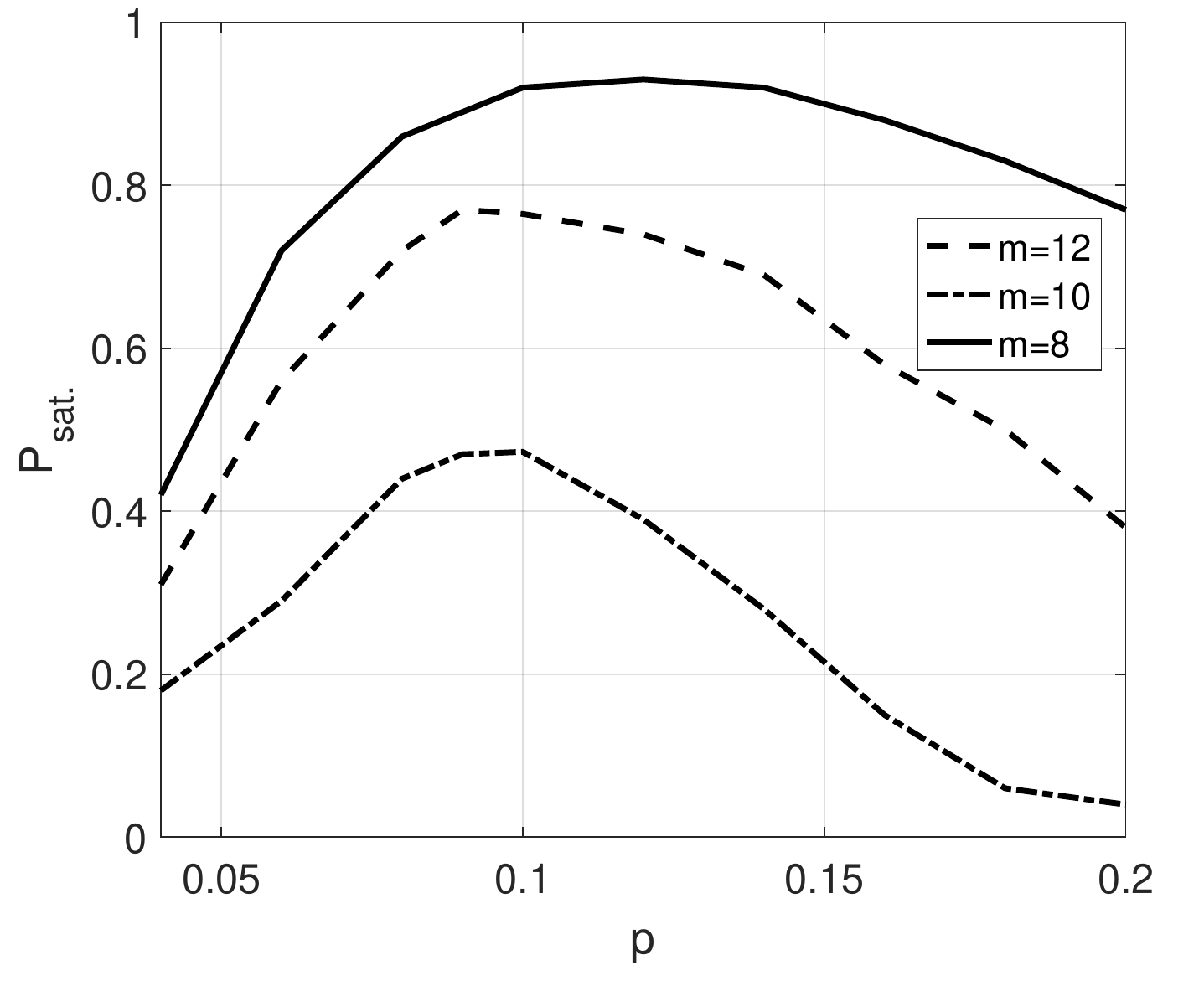}\label{figPsat}}
\end{tabular}
\end{tabularx}
\caption{Average error in computing linear measurements.}
\label{fig6}
\vspace{-5mm}
\end{figure*}

Figure \ref{z1} illustrates a simple network that satisfies Assumption \ref{ass1} in which the unique parents of latent nodes $a, b,c$, and $d$ are $\{1\}$, $\{3\}$, $\{2\}$, and $\{4\}$, respectively. The unique children of latent leaves $c$ and $d$ are $\{5\}$ and $\{2,4\}$, respectively.

\begin{Theorem}\label{theorem2}
Among all unobserved networks that are consistent with the linear measurements induced from (\ref{eq:model}), any graph $G$ that satisfies Assumption \ref{ass1} has the minimum number of latent nodes.
\end{Theorem}

\begin{algorithm}[t]
\begin{algorithmic}[1]
\STATE {\textbf{Input:} $\{Supp(A^*_k)\}_{k\geq 1}$}
\STATE {Find $\{l_i\}$ using (\ref{ew1})} and set $U:=\emptyset$.
\STATE {Find $R_i, M_i$ from (\ref{ew2})} for all $1\leq i\leq n$.
\FOR {$i=1,...,n$}
\STATE  \begin{footnotesize}$Y_i:=\{j: j\neq i \wedge l_j=l_i\}$  \end{footnotesize}
\IF{ \begin{footnotesize}$\forall j\in Y_i$, $(R_j\not\subseteq R_i$) $\vee$ ($R_j=R_i\ \wedge M_i\subseteq M_j$)\end{footnotesize}}
\IF{ $i=\min\{k:R_k=R_i \wedge M_k=M_i\}$}
\STATE{\begin{footnotesize}Create node $h_i$ and set $\mathcal{P}_{h_i}\!\!\!=\!\!\{i\}$, $U\!\!\leftarrow\! \{i\}\cup\! U$ \end{footnotesize}}
\ENDIF

\ENDIF
\ENDFOR
\FOR {every latent node $h_s$ }
\IF{\begin{footnotesize}$\exists h_k, (l_{k}=l_{s}+1) \wedge (R_s\subseteq R_k)$\end{footnotesize}}
\STATE $\mathcal{P}_{h_s}\leftarrow \{h_k\}\cup\mathcal{P}_{h_s}$
\ENDIF 
\STATE {\begin{footnotesize}$\mathcal{C}_{h_s}\leftarrow \{j: [A^*_1]_{js}\neq0\}$\end{footnotesize}}
\ENDFOR
\FOR{$i=1,...,n$}
\IF{\begin{footnotesize}$\exists\ j\in U$, s.t. $M_j\subseteq M_i$ \end{footnotesize}}
\STATE $\mathcal{P}_{h_j}\leftarrow \{i\}\cup\mathcal{P}_{h_j}$
\ENDIF
\ENDFOR
 \caption{DTR Algorithm}
 \label{AlgDTR}
\end{algorithmic}
\end{algorithm}

Note that if Assumption \ref{ass1} is violated, one can find many unobserved networks that are consistent with the linear measurements but are not minimum (in terms of the number of latent nodes).
For example, the network in Figure \ref{v1} satisfies Assumption \ref{ass1} (ii) but not (i). 
Figure \ref{v2} depicts an alternative network with the same linear measurements as the network in Figure \ref{v1} but it has fewer number of latent nodes.
Similarly, the graph in Figure \ref{w1} satisfies Assumption \ref{ass1} (i) but not (ii). Figure \ref{w2} shows an alternative graph with one less latent node.


\begin{Theorem}\label{theorem3}
Consider an unobserved network $G$ with adjacency matrix $Supp([0, A_{12}; A_{21}, A_{22}])$. 
If $G$ satisfies Assumption \ref{ass1}, then its corresponding linear measurements uniquely identify $G$ upto $Supp([0, A_{12}; \widehat{A}_{21}, A_{22}])$, where $Supp(A_{21})\subseteq Supp(\widehat{A}_{21})$.
\end{Theorem}

Figure \ref{z1} gives an example of a network satisfying Assumption \ref{ass1} and an alternative network, Figure \ref{z2}, with the same linear measurements which departs from the Figure \ref{z1} in the $A_{21}$ component.

Next, we propose the directed tree recovery (DTR) algorithm that takes the linear measurements of an unobserved network $G$ satisfying Assumption \ref{ass1} and recovers $G$ upto the limitation in Theorem \ref{theorem3}. This algorithm consists of three main loops. 
Recall that Assumption \ref{ass1} implies that each latent node has at least one unique observed parent.
The first loop finds all the unique observed parents for each latent node (lines: 4-11). 
The second loop reconstructs $Supp(A_{22})$ and $Supp(A_{12})$ (lines: 12-17). And finally, the third loop constructs $Supp(\widehat{A}_{21})$ such that $Supp(A_{21})\subseteq Supp(\widehat{A}_{21})$ (lines: 18-22).

The following lemma shows that the first loop of Algorithm \ref{AlgDTR} can find all the unique observed parents from each latent node. To present the lemma, we need the following definitions.

\begin{Definition}
For an observed node $i$, we define
\begin{small}
\begin{align} \label{ew1}
& l_i:=\max\{k:[A^*_{k-1}]_{si}\neq0, \ \text{for some}\ s\},\\ \label{ew2} 
& R_i:=\{j:[A^*_{l_i-1}]_{ji}\neq0\},\  M_i:=\{(j,r) : [A^*_{r-1}]_{ji}\neq0\}.
\end{align}
\end{small}
\end{Definition}
In the above equations, $l_i$ denotes the length of longest directed latent path that connects node $i$ to any observed node. $R_i$ is the set of all observed nodes that can be reached by $i$ with a directed latent path of length $l_i$ and set $M_i$ consists of all pairs $(j,r)$ such that there exists a directed latent path from $i$ to $j$ with length $r$.

\begin{Lemma}\label{lem}
Under Assumption \ref{ass1}, an observed node $i$ is the unique parent of a latent node if and only if for any other observed node $j$ s.t. $l_{i}=l_{j}$, we have 
$
(R_{j}\not\subseteq R_i) \vee (R_{j}=R_i\ \wedge M_i\subseteq M_j).
$
\end{Lemma}

In the first loop, if there exist multiple unique parents of a latent node (for instance, node 2 and node 3 in Figure \ref{v2}), we pick the one with a minimum index (lines: 7-9). 

 The second loop recovers $Supp(A_{22})$ based on the following observation. If a latent node $h_k$ is the parent of latent node $h_s$, then $h_k$ can reach all the observed nodes in $R_s$, i.e., $R_s\subseteq R_k$  and $l_k=l_s+1$ (line: 13).
 Furthermore, $Supp(A_{12})$ can be recovered using the fact that an observed node $j$ is a children of a latent node $h_s$, if a unique parent of $h_s$, e.g., $s$, can reach $j$ by a directed latent path of length 2 (line: 16).
 Finally, the third loop reconstructs $Supp(\widehat{A}_{21})$ by adding an observed node $i$ to the parent set of latent node $h_j$, if $i$ can reach all the observed nodes that a unique parent of $h_j$, e.g., $j$, reaches (lines: 18-22).

\begin{proposition}\label{prop3}
Suppose network $G$ satisfies Assumption \ref{ass1}. Then given its corresponding linear measurements, Algorithm \ref{AlgDTR} recovers $G$ upto the limitation in Theorem \ref{theorem3}. 
\end{proposition}

\vspace{-3mm}
\subsection{ Learning More General Unobserved Networks with Minimum Number of Latent Nodes}\label{sec:fg}
In general, the latent sub-network may not be a tree or there may not be a unique minimal unobserved network consistent with the linear measurements (see Figure \ref{Fig1}). 
Hence, we try to find an efficient approach to recovering all possible minimal unobserved networks under some conditions.
In fact, without any extra conditions, finding a minimal unobserved network is NP-hard.

\begin{Theorem}\label{theo1}
Finding an unobserved network that is both consistent with a given linear measurements and has a minimum number of latent nodes is NP-hard.
\end{Theorem}
Below, after some definitions, we propose the Node-Merging (NM) algorithm that returns all possible unobserved networks with minimum number of latent nodes under the following assumption. 

\begin{assumption}\label{ass2}
Assume that there exists at most one directed latent path of each length between any two observed nodes.
\end{assumption}

\begin{algorithm}[t]
\begin{algorithmic}[1]
 \STATE {\bf Initialization:} Construct graph $G_0$.
\STATE $\mathcal{G}_0:=G_0$, $\mathcal{G}_s:=\emptyset, \forall s>0$
 \STATE $k:=0$
\WHILE{$\mathcal{G}_k\neq \emptyset$}
\FOR{$ G\in\mathcal{G}_k$}
\FOR{ $i^{\prime},j^{\prime}\in G$}
\IF{$\text{Check}(G,i^{\prime},j^{\prime})$}
\STATE $\mathcal{G}_{k+1}:=\mathcal{G}_{k+1}\cup \text{Merge}(G,i^{\prime},j^{\prime})$.
\ENDIF
\ENDFOR
\ENDFOR
  \STATE $k:=k+1$
 \ENDWHILE
\STATE {\bf Output:} $\mathcal{G}_{out}:=\mathcal{G}_{k-1}$
 \caption{NM Algorithm}
 \label{AlgNM}
\end{algorithmic}
\end{algorithm}

For example, the graph in Figure \ref{z2} satisfies this assumption but not the one in Figure \ref{z1}. This is because there are two directed latent paths of length 2 from node 5 to node 4.

\begin{Definition}
(Merging)
We define merging two nodes $i'$ and $j'$ in graph $G$ as follows: remove node $j^{\prime}$ and the edges between $i'$ and $j'$, and then give all the parents and children of $j'$to $i^{\prime}$. 
We denote the resulting graph after merging $i'$ and $j'$ by $\text{Merge}(G,i',j')$.
We say that two nodes $i^{\prime}$ and $j^{\prime}$ are mergeable if $\text{Merge}(G,i',j')$ is consistent with the linear measurements of $G$.
\end{Definition}

\begin{Definition} (Connectedness) Consider an undirected graph $\bar{G}$ over the observed nodes which is constructed as follows: there is an edge between two nodes $i$ and $j$ in $\bar{G}$, if there exists $ k\geq 1$ s.t. $ Supp([A_k^*]_{ij})=1$ or $Supp([A_k^*]_{ji})=1$;
We say that two observed nodes $i$ and $j$ are ``connected" if there exist a path between them in $\bar{G}$.
\end{Definition}

It can be seen that if pairs $i,j$ and $j,k$ are connected then node $i,k$ are also connected. 
We then define a \textit{connected class} as a subset of observed nodes in which any two nodes  are connected.

{\bf Initialization:}
We first find the set of all connected classes, say $S_1,S_2,...,S_C$. 
For each class $S_c$, we create a directed graph $G_{0,c}$ that is consistent with the linear measurements. 
To do so, for any two observed nodes $i,j\in S_c$, if $[A_r^*]_{ji}\neq 0$, we construct a directed path with length $r+1$ from node $i$ to node $j$ by adding $r$ new latent nodes to $G_{0,c}$. 
\\
{\bf Merger:}
In this phase, for any $G_{0,c}$ from the initialization phase, we merge its latent nodes iteratively until no further latent pairs can be merged. 
Since the order of mergers leads to different networks with minimum number of latent nodes, the output of this phase will be the set of all such networks.
Algorithm \ref{AlgNM} summarizes the steps of NM algorithm.
In this algorithm, subroutine $\text{Check}(G,i',j')$ checks whether two nodes $i'$ and $j'$ are mergeable.

\begin{Theorem}\label{NMth}
Under Assumptions \ref{ass0} and \ref{ass2}, the NM algorithm returns the set of all networks that are consistent with the linear measurements and have minimum number of latent nodes.
\end{Theorem}

\vspace{-5mm}
\section{Experimental Results}\label{sec:exp}
\vspace{-1mm}
\paragraph{Synthetic Data:} We considered a directed random graph, denoted by DRG$(p,q)$, such that there exists a directed link between an observed and latent node with probability $p$,  independently across all pairs, and there is a directed link between two latent nodes with probability $q$. If there is a link between two nodes, we set the weight of that link uniformly from $[-a,a]$.

We utilize the method described in Section 3 to estimate linear measurements with a significance level of $0.05$.
In order to evaluate how well we can estimate the linear measurements, we generated 1000 instances of DRG$(0.4,0.4)$ with $n+m=100$, $\Sigma_X=0.1I_{n\times n},\allowbreak\Sigma_Z=0.1I_{m \times m}$, and $a=0.1$. 
The length of the time series was set to $T=1000$. Let $Supp(\hat{A}_{11})$ be the estimate of support of $A_{11}$.
In Figure \ref{fig6a}, the expected estimation error, i.e. $|| Supp(\hat{A}_{11})-Supp(A_{11})||^2_F/n^2$, is computed, where $||.||_F$ is the Frobenius norm. One can see that the
estimation error decreases as the number of observed variables increases.

\begin{figure}[t]
\def\tabularxcolumn#1{m{#1}}
\hspace{0cm}
\begin{tabularx}{\linewidth}{@{}cXX@{}}
\begin{tabular}{cc}
\subfloat[The percentage of instances that can be reconstructed efficiently in time.]{\hspace{0cm}\includegraphics[width=1.68in,height=1.3in]{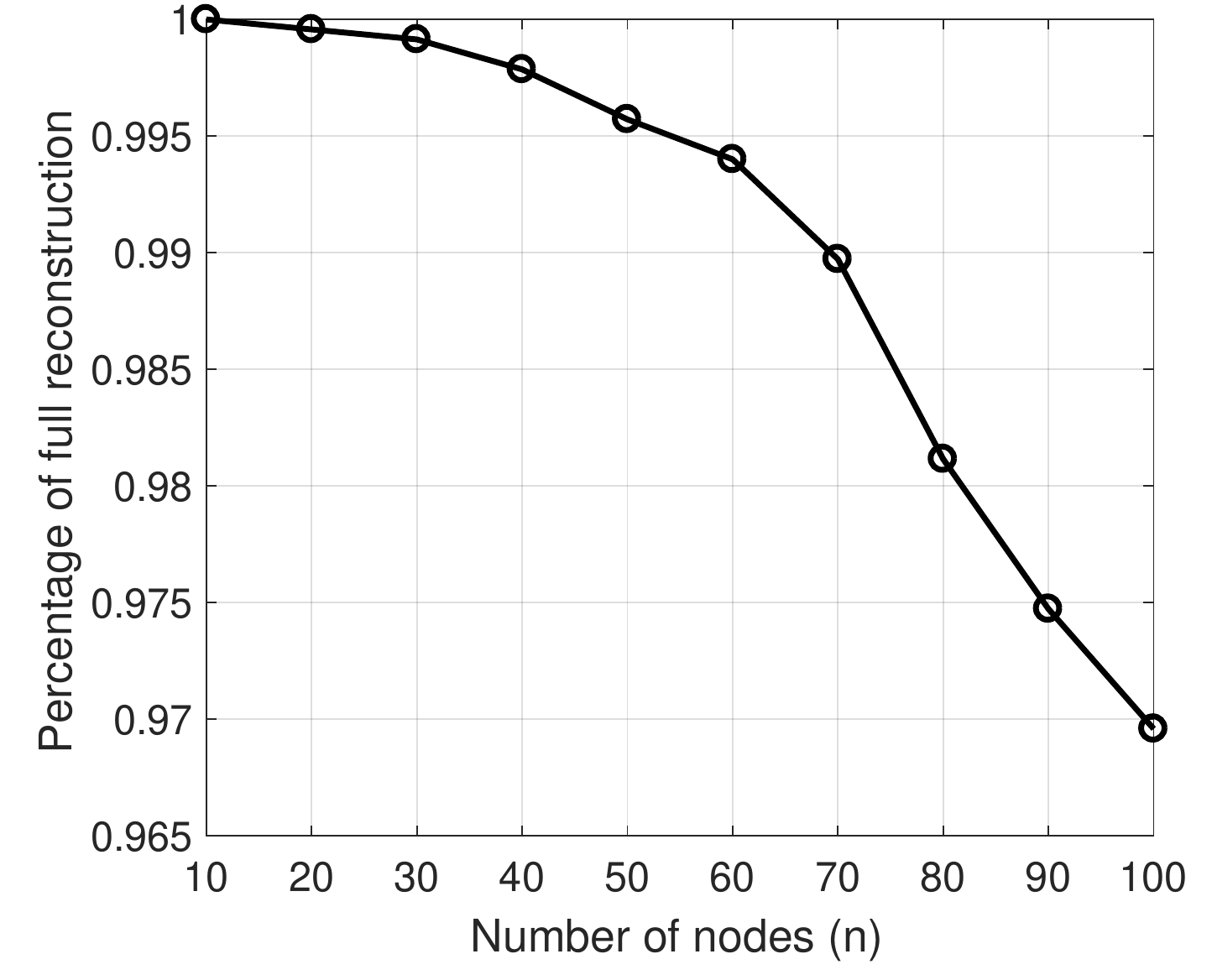}\label{fig5a}}
&
\subfloat[Average run time of the algorithm.]{\hspace{-.4cm}\includegraphics[width=1.68in,height=1.3in]{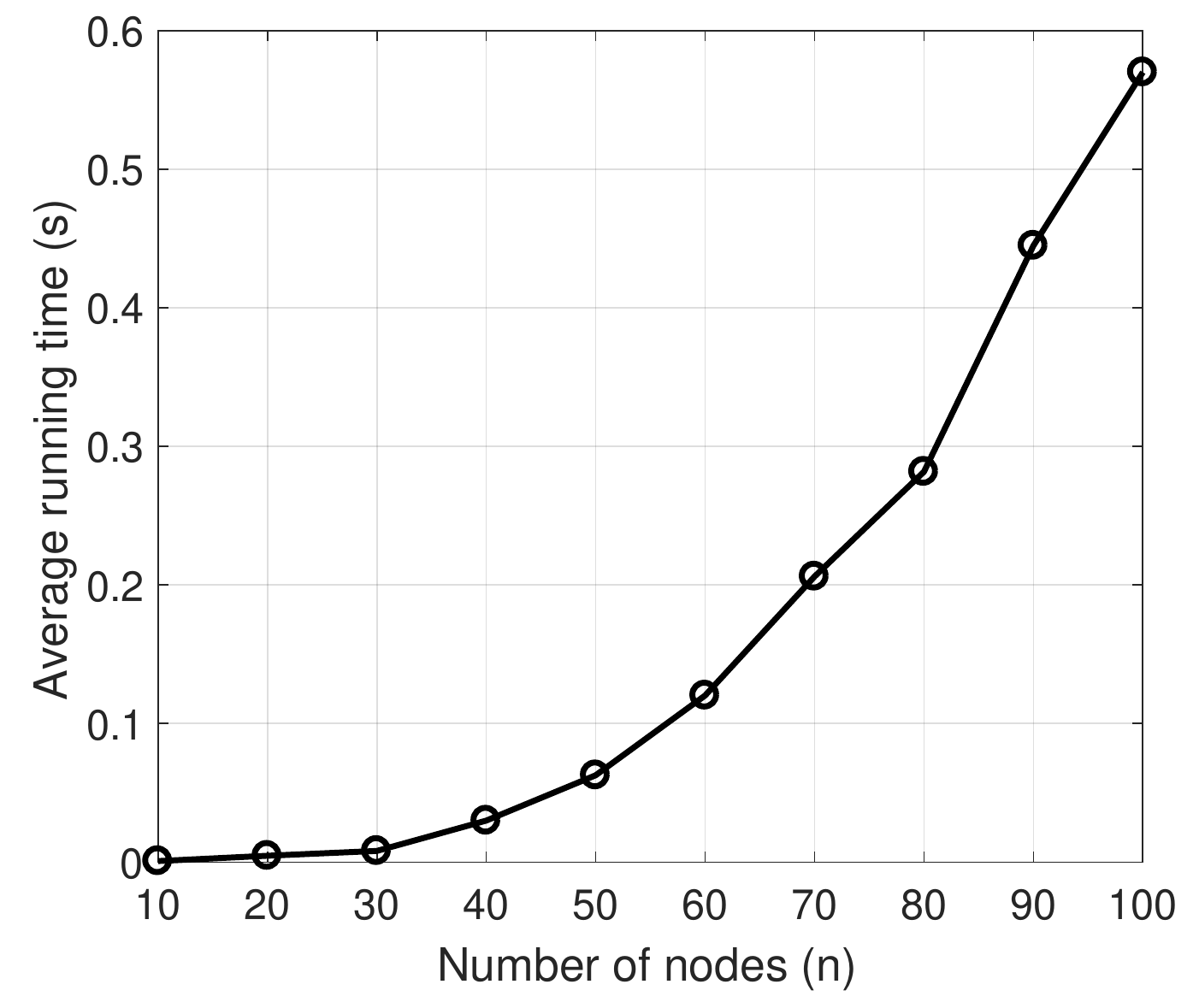}\label{fig5b}}
\end{tabular}
\end{tabularx}
\caption{Recovering the minimal unobserved network for instances of DRG$(1/(2n),1/(2n))$ where $n\in\{10,...,100\}$, $m=n/2$.}\label{fig:f}
\vspace{-5mm}
\end{figure}

We also studied the effect of the observed to latent noise power ratio (OLNR), $\sigma_X^2/\sigma_Z^2$, on $||B_0^*-A_0^*||_1$, and compared it with the bound given in Proposition \ref{prop1}. We generated $1000$ instances of DRG$(0.05,0.05)$ with $n=5$, $m=5$, and $a=0.1$. As it can be seen in Figure \ref{fig6b}, the average estimation error decreases as OLNR increases, as expected from Proposition \ref{propSNR}.

We investigated what percentage of instances of the random graphs satisfy Assumption \ref{ass1}. 
We generated $1000$ instances of DRG$(p,1/n)$ with $n=100$, and $p\in[0.04,0.2]$.
 In Figure \ref{figPsat}, the probability of satisfying Assumption \ref{ass1}, $P_{sat.}$, is depicted versus  $p$ for different numbers of latent variables in the VAR model. 
 For larger $m$, it is less likely to see a unique observed parent for each latent node and thus $P_{sat.}$ decreases.
 For a fixed $m$, the same phenomenon will occur if we increase $p$ when $p$ is relatively large. 
 Furthermore, for small $p$, there might exist some latent nodes that have no observed parent or no observed children.

We also evaluated the performance of the NM algorithm in random graphs. 
We generated $1000$ instances of DRG$(1/2n,1/2n)$ with $n\!=\!10,...,100$ and $m\!=\!n/2$, and computed the linear measurements. 
To save time, if for a class of connected nodes the number of latent nodes generated in the initial phase exceeds $40$, 
we supposed that the corresponding instance cannot be recovered efficiently in time and did not proceed to the merging phase. 
Figures \ref{fig5a} and \ref{fig5b} depict the percentage of instances in which the algorithm can recover all possible minimal unobserved networks and the average run time (in seconds) of the algorithm, respectively.\footnote{We performed the experiment on a Mac with $2\!\times\! 2.4$ GHz 6-Core Intel Xeon processor and 32 GB of RAM.} 
This plot shows that we can recover all possible minimal unobserved networks for a large portion of instances efficiently even in relatively large networks.

\begin{figure}[t]
\def\tabularxcolumn#1{m{#1}}
\hspace{0cm}
\begin{tabularx}{\linewidth}{@{}cXX@{}}
\begin{tabular}{cc}
\subfloat[High power]{\hspace{-.5cm}\includegraphics[width=1.7in,height=1.4in]{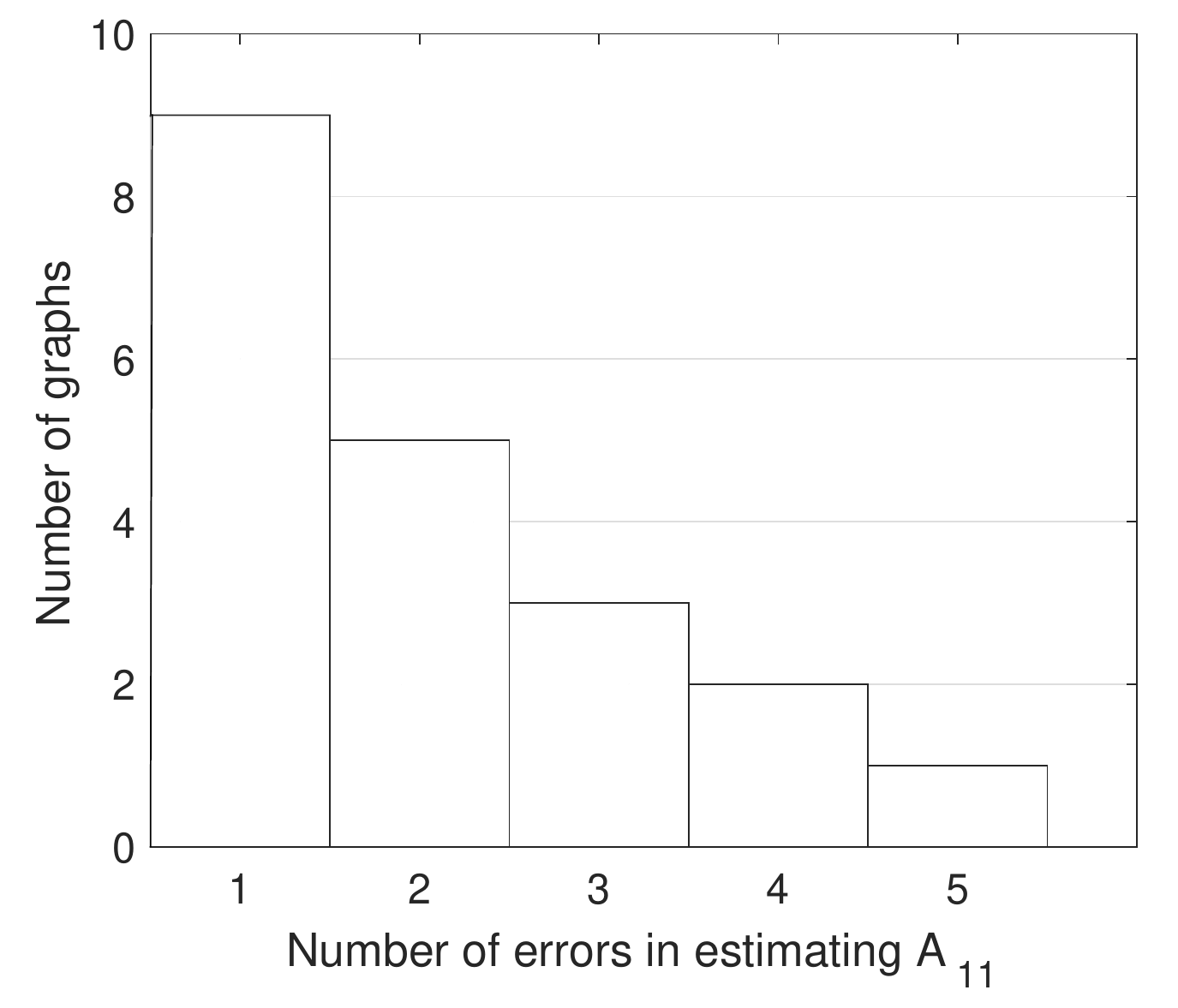}\label{fig7a}}
&
\subfloat[Low power]{\hspace{-.5cm}\includegraphics[width=1.7in,height=1.4in]{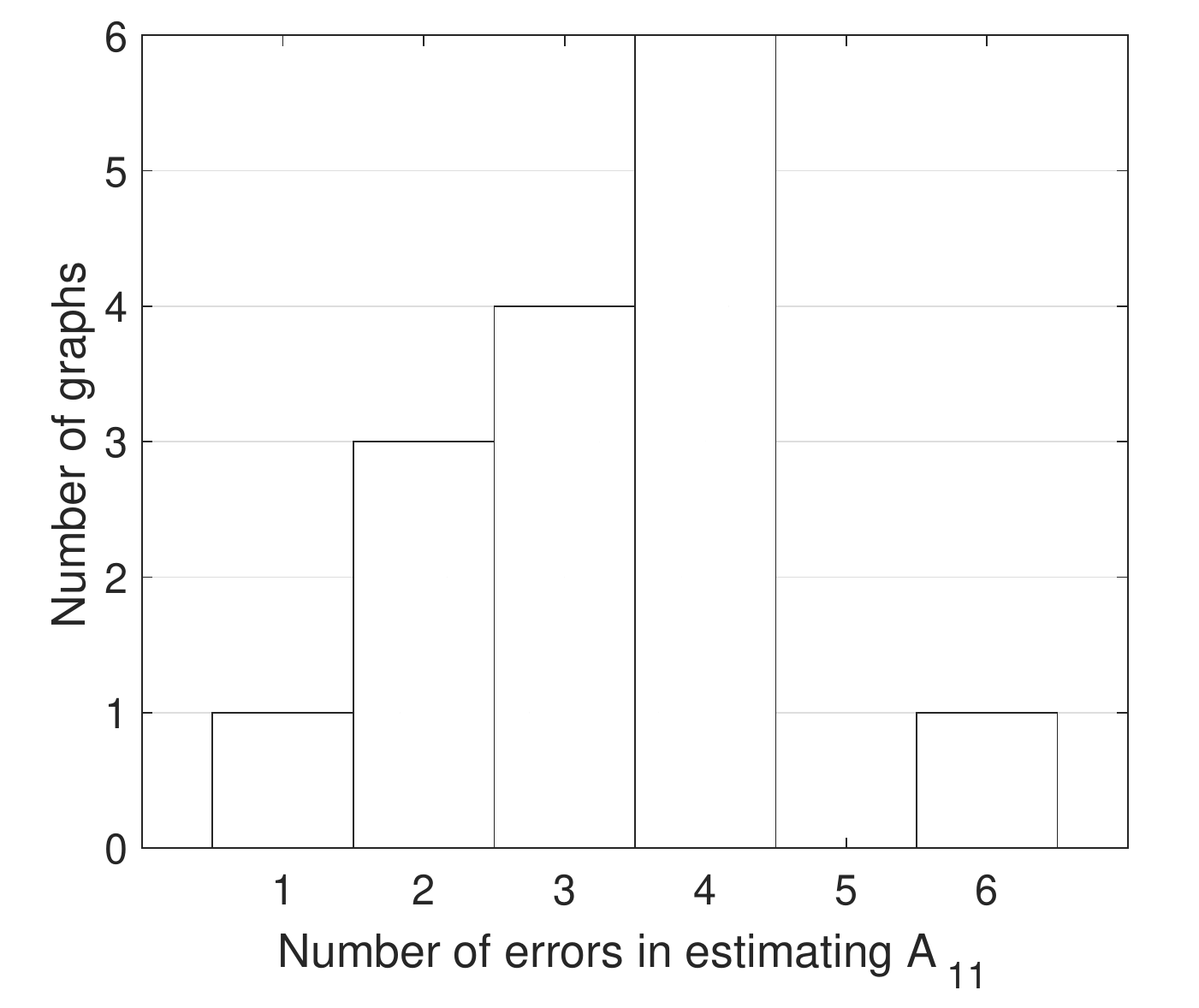}\label{fig7b}}
\end{tabular}
\end{tabularx}
\vspace{-3mm}
\caption{Histogram of $|| Supp(\hat{A}_{11})-Supp(A_{11})||^2_F$.}\label{fig:f}
\vspace{-7mm}
\end{figure}

\vspace{-.5cm}
\paragraph{US Macroeconomic Data:}
We considered the following set of time series from the quarterly US macroeconomic data for the period from 31-Mar-1947 to 31-Mar-2009 collected from the St. Louis Federal Reserve Economic Database (FRED) \cite{FRED}:
GDP, GDPDEF, COE, HOANBS, TB3MS, PCEC, GPDI.

Assuming that the underlying dynamics is linear (Eq. \eqref{eq:model}), we considered the estimated VAR model over all variables as the ground truth.
Then, we selected four arbitrary times series as observed processes and computed $Supp(\hat{A}_{11})$. 
We divided the ${7 \choose 4}=35$ possible selections into two classes: 1) high power, where tr$(\mathbb{E}\{\omega_X(t)\omega_X(t)^T\})>\tau$ for a fixed threshold $\tau$; 
2) low power:  where tr$(\mathbb{E}\{\omega_X(t)\omega_X(t)^T\})<\tau$.
In this experiment, we set $\tau=0.02$.
 In Figure \ref{fig:f}, we plotted the histograms of $|| Supp(\hat{A}_{11})-Supp(A_{11})||^2_F$ for these two classes. 
As it can be seen, in the high power regime, most of the possible selections have small estimation errors.

\begin{wrapfigure}{r}{5cm}
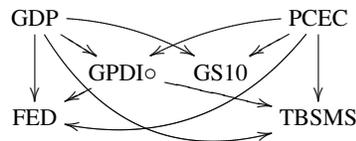

{\xygraph{ !{<0cm,0cm>;<1.3cm,0cm>:<0cm,1.2cm>::} 
!{(-.9,.4) }*+{\footnotesize{\text{GDP}}}="gdp"   
  !{(0,-.2) }*+{\footnotesize{\text{GPDI}}\circ}="gpdi"	
	!{(2,.4) }*+{\footnotesize{\text{PCEC}}}="pcec"   	 		
!{(1,-.2) }*+{\footnotesize{\text{GS10}}}="gsio" 
!{(2,-.7) }*+{\footnotesize{\text{TBSMS}}}="tbsms"  
!{(-.9,-.7) }*+{\footnotesize{\text{FED}}}="fed" 
"gdp":@/^/"gsio" "gdp":"fed" "gdp":@/_{.9cm}/"tbsms"
"gdp":"gpdi" "pcec":@/_/"gpdi" "pcec":"gsio" "pcec":"tbsms" "gpdi":"fed" "gpdi":"tbsms" "pcec":@/_{-.7cm}/"fed"
} }
\caption{US macroeconomic data.}\label{FDGUS}
\vspace{-.5cm}
\end{wrapfigure}

We also considered the following six time series of US macroeconomic data during 1-Jun-2009 to 31-Dec-2016 from the same database: GDP, GPDI, PCEC, TBSMS, FEDFUND, and GS10.  We obtained the causal structure among these six time series by fitting a VAR model on all of them and considered the result as our ground truth (see Figure \ref{FDGUS}). 
Then, we removed GPDI from the dataset and considered the remaining five time series as observe processes and checked whether the influences from the ``latent" process (GPDI) can be corrected estimated. 
We estimated the linear measurements and gave them as an input  to Algorithm \ref{AlgDTR}, which successfully recovered the ground truth (the estimated structure, in which the latent process is denoted by a circle, is identical to that in Figure \ref{FDGUS}).

\vspace{-5.5mm}
\paragraph{Dairy Prices:}
A collection of three US dairy prices has been observed monthly from January 1986 to December 2016 \cite{Dairy}: milk, butter, and cheese prices. 
\begin{wrapfigure}{r}{4.7cm}
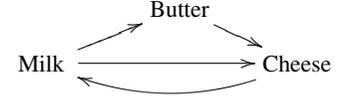

{\xygraph{ !{<0cm,0cm>;<1.2cm,0cm>:<0cm,.8cm>::} 
!{(-.5,-.6) }*+{\small{\text{Milk }}}="x1"   
  !{(1,.3) }*+{\small{\text{Butter} }}="z"	
   !{(2.3,-.6) }*+{\small{\text{Cheese}}}="x2"  
"x1":"x2" "x1":"z"  "z":"x2"  "x2":@/_{-.4cm}/"x1"
} }
\caption{Dairy prices}\label{DairyDFG}
	\vspace{-.55cm}
\end{wrapfigure} 
We estimated the VAR model on all the time series with lag length $l\!=\!1$  and considered the resulting graph as our ground truth (see Figure \ref{DairyDFG}).
 Next, we omitted  the butter prices from the dataset and considered  
 the milk and cheese prices as observed processes.
 The estimated linear measurements were: $Supp(A_0^*)\!=\!Supp(A_{11})\!=\![1,1;1,0]$ and $Supp(A_1^*)\!=\![0,0;1,0]$. 
Algorithm \ref{AlgDTR} correctly recovered  the true causal graph using this linear measurements. Note that the genericity assumptions in \cite{Geiger} do not hold true for this data set (see Experiments section).

\vspace{-5mm}
\paragraph{West German Macroeconomic Data:}We considered the quarterly West German consumption expenditures $X_1$, fixed investment $X_2$, and disposable income $X_3$, during 1960-1982 \cite{WG}. 
\begin{wrapfigure}{r}{4.7cm}
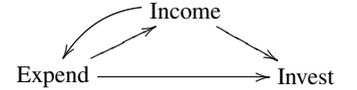

\vspace{-.1cm}
\hspace{.3cm}
{\xygraph{ !{<0cm,0cm>;<1.2cm,0cm>:<0cm,.8cm>::} 
!{(-.5,-.6) }*+{\small{\text{Expend}}}="x1"   
  !{(1,.5) }*+{\small{\text{Income }}}="z"	
   !{(2.3,-.6) }*+{\small{\text{Invest}}}="x2"  
"x1":"x2" "x1":"z"  "z":"x2"  "z":@/_{.4cm}/"x1"
} }
\caption{West German macroeconomic data.}\label{figWG}
	\vspace{-.5cm}
\end{wrapfigure} 
Similar to the previous experiment with dairy prices, we first obtained the entire transition matrix among all the process. Figure \ref{figWG} depicts the resulting graph.
Next, we considered $X_3$ to be latent and used $\{X_1,X_2\}$ to estimate the linear measurements $Supp(A_0^*)\!=\!Supp(A_{11})\!=\![0,0;1,1]$ and $Supp(A_1^*)\!=\![1,0;1,0]$. Using this linear measurements, Algorithm \ref{AlgDTR} recovered the true network in Figure \ref{figWG} correctly.

\vspace{-3mm}
\section{Conclusion and Future work}

We considered the problem of estimating time-delayed influence structure from partially observed time series data. 
Our approach consisted of two parts: First, we studied sufficient conditions under which certain aspects of the influence structure of the underlying system are identifiable. Second, we proposed two algorithms that recover the influence structures satisfying the sufficient conditions given in the first part. The proposed algorithms can construct the observed sub-network (support of $A_{11}$), the causal influences from latent to observed processes (support of $A_{12}$), and also the causal influences among the latent variables (support of $A_{22}$), uniquely under a set of sufficient conditions. As a future direction, we plan to extend our results to the case that $A_{22}$ might have cycles. In the paper, we have seen examples showing that unique recovery is not possible if any conditions of Assumption \ref{ass1} are violated. These conditions are a good starting point for the case that we have cycles in $A_{22}$.
\bibliography{Bibliography-File}
\bibliographystyle{aaai}

\newpage
\newpage
\appendix

\section{Proof of Proposition \ref{prop1}}\label{pprop1}
We project the vector $\widetilde{A}_{r+1:l-1}[\vec{\omega}_Z(t-r-1);\cdots;\vec{\omega}_Z(t-l+1)]$ onto $\vec{X}(t-r)$ as follows: 
\begin{small}
\begin{align}
\widetilde{A}_{r+1:l-1}\begin{bmatrix}
\vec{\omega}_Z(t-r-1)\\
\vdots\\
\vec{\omega}_Z(t\!-\!l\!+\!1)
\end{bmatrix}=\textbf{C}_r
\vec{X}(t-r)
+\begin{bmatrix}
\vec{N}_Z(t-r-1)\\
\vdots\\
\vec{N}_Z(t\!-\!l\!+\!1)
\end{bmatrix},
\end{align}
\end{small}
where $\widetilde{A}_{r+1:l-1}=diag(\tilde{A}_{r+1},...,\tilde{A}_{l-1})$, and $\textbf{C}_r$ is a block matrix with $C_r^s$ as its $s$th block for $s=0,...,l-\!r-\!2$. Please note that $\vec{\omega}_Z(t-r)$ is orthogonal to $\vec{X}(t-k)$ for $k\geq r$. 
Since $\vec{N}_{Z}$ and $\vec{X}(t-r)$ are orthogonal, we can see 
\begin{align}\label{eq:ine}
||\widetilde{A}_{r+1:l-1}\Gamma_{\omega_Z}(l\!-r\!-\!2)\widetilde{A}_{r+1:l-1}^T||_2\geq ||\textbf{C}_r\Gamma_{X}(0)\textbf{C}_r^T||_2.
\end{align}
Using (\ref{eq:ine}) and the relationship between $\ell_2$ and $\ell_1$  norms of a matrix, we obtain
\begin{align}\label{eq:in2}
\nonumber\lambda_{max}\left(\Gamma_{\omega_Z}(0)\right)||\widetilde{A}&_{r+1:l-1}||_2^2\geq \\
&\lambda_{min}\left(\Gamma_{X}(0)\right)||\textbf{C}_r||_1^2/(n(l\!-\!r\!-\!1)),
\end{align}
 where $\lambda_{min}(\cdot)$ and $\lambda_{max}(\cdot)$ denote the minimum and maximum eigenvalues of a given matrix, respectively. Please note that $\vec{\omega}_Z(t)$ is white noise and thus we have: $\lambda_{max}(\Gamma_{\omega_Z}(l-r-2))=\lambda_{max}(\Gamma_{\omega_Z}(0))$. Using the fact that $\widetilde{A}_{r+1:l-1}$ is diagonal and $||A_{22}||_2<1$, we obtain
\begin{small}
\begin{align}\label{eq:finnew}
\nonumber ||\textbf{C}_r||_1 &\leq \sqrt{n(l\!-\!r\!-\!1)\frac{M}{L}}||A_{12}||_2\!\!\!\max_{r+1\leq k\leq l-1}||A_{22}||^k_2\\
&\leq  \sqrt{n(l\!-\!r\!-\!1)\frac{M}{L}}||A_{12}||_2||A_{22}||_2^{r+1}.
\end{align}
\end{small}
where $M:=\lambda_{max}(\Gamma_{\omega_Z}(0))$ and $L:=\lambda_{min}\left(\Gamma_{X}(0)\right)$.

From (\ref{matrixform}), we have $B^*_r-A_r^*= \sum_{s=0}^{l-r-2} C_r^s$. This implies that 
$||B^*_r-A_r^*||_1\leq ||\textbf{C}_r||_1$. Combining this inequality and the bound in (\ref{eq:finnew}) concludes the result.

\section{Estimating the Bounds in Proposition \ref{prop1}}\label{bounds}
The bound $\sqrt{n(l\!-\!r\!-\!1)\frac{M}{L}}||A_{12}||_2||A_{22}||_2^{k+1}$ can be estimated as follows:
\begin{itemize}
\item The lag length $l$ in \eqref{matrixform} can be obtained from AIC or FPE criterion (see chapter 4 in \cite{lutkepohl2005new}).
\item We can estimate $L$ by observation vector $\vec{X}(t)$. We also consider a bound $\sigma^2_{max,Z}$ on the maximum variance of exogenous noises in latent part.
\item We assume a bound on $||A_{12}||_2\leq \rho_{12}$ and $||A_{22}||_2\leq \rho_{22}<1$. 
\end{itemize}

In summary, an upper bound would be: $\sqrt{n(l\!-\!r\!-\!1)\frac{\sigma^2_{max,Z}}{L}} \rho_{12} \rho_{22}^{k+1}$. Suppose that absolute values of nonzero entries of $A_k^*$ are greater than $a_{min,k}$. We can recover the support of matrix $A_k^*$ successfully if
\begin{align}
\frac{4n(l\!-\!r\!-\!1)\rho_{12}^2}{a_{min,k}^2}(\rho_{22})^{2(k+1)}\leq \frac{L}{\sigma_{max,Z}^2}.
\end{align}

 \section{Proof of Proposition \ref{propSNR}}\label{propSNRA}
The spectral density of matrix $\gamma_X(h)$ can be computed as follows:
\begin{equation}
\mathcal{F}(\gamma_X)= \sigma_X^2 F_X(\Omega) F_X(\Omega)^{H}+ \sigma_Z^2 F_Z(\Omega) F_Z(\Omega)^{H}
\end{equation}
where $F_X(\Omega)= [e^{j\Omega}I_{n\times n} -A_{11} -\sum_{k=0}^{l-1} A_k^*e^{-kj\Omega}]^{-1}$, $F_Z(\Omega)=F_X(\Omega)(A_{12}\sum_{k=0}^{l-1} A_{22}^k \allowbreak \times e^{-kj\Omega})$, and $H$ denotes Hermitian of a matrix. Thus, we have:
\begin{align}
\Gamma_X(0)=\frac{1}{2\pi} \int_0^{2 \pi} \mathcal{F}(\gamma_X) d\Omega= \sigma_X^2 F_X^0+\sigma_Z^2 F_Z^0,
\end{align}
where $F_X^0=1/(2\pi)\int_0^{2\pi} F_X(\Omega) F_X(\Omega)^{H} d\Omega$ and $F_Z^0=1/(2\pi)\int_0^{2\pi} F_Z(\Omega) F_Z(\Omega)^{H}d\Omega$.

We define the function $\psi_{\frac{\sigma_X}{\sigma_Z}}(v):=\vec{v}^T\Gamma_X(0)\vec{v}/\sigma_Z^2= (\sigma_X^2/\sigma_Z^2)F_X^0+ F_Z^0$ where $\vec{v}$ is a unit vector. Suppose that $\vec{v}^*$ minimizes the function $\psi_{\frac{\sigma_X}{\sigma_Z}}(.)$. By the definition of $L$ and $M$, the ratio $M/L$ is equal to $1/\psi_{\frac{\sigma_X}{\sigma_Z}}(\vec{v}^*)$. Now if we decrease $\frac{\sigma_X}{\sigma_Z}$ to $\frac{\sigma'_X}{\sigma'_Z}$, then we have: $\psi_{\frac{\sigma'_X}{\sigma'_Z}}(\vec{v}^*)<\psi_{\frac{\sigma_X}{\sigma_Z}}(\vec{v}^*)$. Moreover, for the optimal solution $\vec{v}'^*$ of $\psi_{\frac{\sigma'_X}{\sigma'_Z}}(.)$, we know that: $\psi_{\frac{\sigma'_X}{\sigma'_Z}}(\vec{v}'^*)\leq\psi_{\frac{\sigma'_X}{\sigma'_Z}}(\vec{v}^*)$. Thus, we can conclude that: $1/\psi_{\frac{\sigma'_X}{\sigma'_Z}}(\vec{v}'^*)>1/\psi_{\frac{\sigma_X}{\sigma_Z}}(\vec{v}^*)$. 

\section{Proof of Theorem \ref{theorem2}}\label{ptheorem2}
First, we show such $G$ has a minimum number of latent nodes. We do this by means of contradiction. 
But first observe that since the latent subnetwork of $G$ is a directed tree, we can assign a non-negative number $l_h$ to latent node $h$ that represents the length of longest directed path from $h$ to its latent descendants. Clearly, all such descendants are leaves which we denote them by $\tilde{L}_h$.
For instance, if the latent subnetwork of $G$ is $a\rightarrow b\rightarrow c$, then $l_a=2$ and $\tilde{L}_a=\{c\}$. 

Suppose that $G$ contains $m$ latent nodes $\{h_1,...,h_m\}$ and there exists another network $G_1$ (not necessary with tree-structure induced latent subgraph), with $m_1<m$ number of latent nodes that it is also consistent with the same linear measurements as $G$.
Due to assumption (i), there is at least $m$ distinct observed nodes that have out-going edges to the latent subnetwork. More precisely, each $h_i$ has at least a unique observed node as its parent. We denote a unique observed parent of node $h_i$ by $o_i$.

Because $m_1<m$, there exists at least one observed node in $\bar{O}:=\{o_1,...,o_m\}$ that has shared its latent children with some other latent nodes in $G_1$. Among all such observed nodes, let $o_{i^*}$ to be the one whose corresponding latent node in $G$, ($h_{i^*}$), has maximum $l_{h_{i^*}}$.\footnote{If there are several such observed node, let $o_{i^*}$ to be one of them.}
Furthermore, let $\tilde{I}_{i^*}\subset \{1,...,m\}\setminus\{i^*\}$ to be the index-set of those observed nodes that $o_{i^*}$ has shared a latent child with them in $G_1$.

By the choice of $o_{i^*}$, we know that $l_{h_j}\leq l_{h_{i^*}}$ for all $j\in \tilde{I}_{i^*}$ and if for some $1\leq k\leq m$, $l_{h_k}>l_{h_{i^*}}$, then $o_k$ has not shared its latent child in $G_1$ with any other observed nodes in $\bar{O}$. Moreover, there should be at least a latent node $h_{j^*}$ where $j^*\in \tilde{I}_{i^*}$ such that $l_{h_{j^*}}= l_{h_{i^*}}$. Otherwise, $G_1$ will not be consistent with the linear measurements of $G$.
Let $\tilde{I}_{**}:=\{j: l_{h_{j}}= l_{h_{i^*}}\}\cap\tilde{I}_{i^*}$. Because $o_{i^*}$ shares its latent children with $\cup_{j\in\tilde{I}_{**}}o_j$ in $G_1$ and both $G$ and $G_1$ consistent with the same linear measurements,  the following holds in graph $G$,
\begin{align*}
&\mathcal{C}^O_{\tilde{L}_{h_{i^*}}}(G)\subseteq \cup_{j\in\tilde{I}_{**}}\mathcal{C}^O_{\tilde{L}_{h_{j}}}(G),
\end{align*}
where $\mathcal{C}^O_{\tilde{L}_{h_{j}}(G)}$ indicates the set of observed children of the set $\tilde{L}_{h_{j}}$. 
This indeed contradicts assumption (ii).

\section{Proof of Theorem \ref{theorem3}}\label{ptheorem3}
First, we require the following definition. 
For a network $G$ with corresponding latent sub-network that is a tree, we define $U_k(G):=\{h\in G: \ l_h=k\}$.
To prove the equivalency, suppose there exists another network $G_2$ such that its latent sub-network is a tree and has a minimum number of latent nodes.
Let $\{h_1,...,h_m\}$ to denote the latent nodes in $G$. 
Since $G$ satisfies Assumption (i), for every latent node $h_i$ there exists a unique observed node $o_i$ such that $o_i\in \mathcal{P}^O_{h_i}(G)$ and $o_j\not\in \mathcal{P}^O_{h_i}(G)$ for all $j\neq i$.

Since both $G$ and $G_2$ are consistent with the same linear measurement, it is easy to observe that if $h_i\in U_k(G)$, then $o_i$ must have at least a latent child in $G_2$, say $h'_i$, such that $l_{h_i}=l_{h'_i}$. Note that $l_{h_i}$ is computed in $G$ and  $l_{h'_i}$ in $G_2$.
Moreover, we must have: 
\begin{align*}
\mathcal{C}_{\tilde{L}_{h_i}}^O(G)=\bigcup_{h'\in H'(o_i)\cap U_{l_{h_i}}(G_2)}\mathcal{C}^O_{\tilde{L}_{h'}}(G_2), \ 
\end{align*}
where $H'(o_i)$ denotes the set of latent nodes in $G_2$ that have $o_i$ as their observed parent.
 In other words, observed nodes that can be reached by a directed path of length $l_{h_i}+2$ from $o_i$ should be the same in both graph $G$ and $G_2$.
This results plus the fact that $G$ satisfies Assumption (ii), imply: 
\\
I) For every $h_i\in U_k(G)$, there exists a unique latent node $h'_i\in U_k(G_2)$, such that $o_i\in \mathcal{P}^O_{h'_i}(G_2)$ and $o_j\not\in \mathcal{P}^O_{h'_i}(G_2)$ for all $j\neq i$, and
\begin{align*}
\mathcal{C}^O_{\tilde{L}_h}(G)=\mathcal{C}_{\tilde{L}_{h'_i}}^O(G_2).
\end{align*}
Using I) and knowing that both $G$ and $G_2$ have the same number of latent nodes, we obtain:\\
II) $|U_k(G)|=|U_k(G_2)|$, for all $k$.
\\
Using I) and II), we can define a bijection $\phi$ between the latent subnetworks of $G$ and $G_2$ as follows $\phi(h_i)=h'_i$. 
Using this bijection and Assumption (ii) of $G$ conclude that if $h\in U_k(G)$ is the common parent of $\{h_{j_1},...,h_{j_s}\}\subseteq U_{k-1}(G)$, then  $\phi(h)\in U_k(G_2)$ should be the common parent of $\{\phi(h_{j_1}),...,\phi(h_{j_s})\}\subseteq U_{k-1}(G_2)$ and the proof is complete.

\section{Proof of Lemma \ref{lem}}\label{plem}
Suppose that $o_i$ is the unique observed node of a latent node $h_i$. Then, for any $o_j$ such that $l_i=l_j$, if $h_i$ is not a child of $o_j$, then from assumption ii we have $R_j\not\subseteq R_i$. If $h_i$ is a child of $o_j$, then since we know that $l_i=l_j$, $M_i\subseteq M_j$ and $R_i=R_j$. 

Now, suppose that the observed node $o_i$ satisfies conditions but it is not unique parent of any latent node. Let $h_i$ and $h_i^{\prime}$ be children of $o_i$. At least one of them, say node $h_i$, can reach an observed node by a path of length $l_i-1$. If $h_i^{\prime}$ has the same property, then consider the unique observed parent of $h_i^{\prime}$, say node $o_j$. Based on Assumption (ii), we have $R_j\subseteq R_i$, which is in contradiction with the assumption that node $o_i$ satisfies conditions of Lemma. Moreover, if $h_i^{\prime}$ does not have a path to observed node with a length of $l_i-1$, then for any observed parent of $h_i$, one of the conditions in the Lemma is not satisfied. Thus, the proof is complete.

\section{Proof of Proposition \ref{prop3}}\label{p:prop3}
Notice that the first loop in Algorithm \ref{AlgDTR} uses the result of Lemma \ref{lem} and finds all the latent nodes and their corresponding unique observed parents.
The next loop uses the fact that the latent sub-network is a tree and also it satisfies Assumption \ref{ass1}. Hence, if there exist two latent nodes $h$ and $h'$, one with depth $l$ and the other one with depth $l+1$, such that $R_h\subseteq R_{h'}$, then $h'$ must be the parent of $h$ in the latent sub-network. 

Moreover, since each latent node has a unique observed parent, using $A^*_1$, Algorithm \ref{AlgDTR} can identify all the observed children of a latent node.
Finally, the last loop in this algorithm locates the rest of observed nodes as the input of the right latent nodes. The algorithm does it by using the fact that if an observed node $i$ shares a latent child with another observed node $j\in U$, then $M_j\subseteq M_i$.
Clearly, if the true unobserved network satisfies Assumption \ref{ass1}, the output of this algorithm will have a latent sub-network that is a tree and consistent with the linear measurement. Thus, by the result of Theorem \ref{theorem2}, it will be the same as the true unobserved network up to some permutations in $Supp(A_{21})$.

\section{Proof of Theorem \ref{theo1}}\label{ptheo1}
Consider the instance of the problem where $A_{22}=0_{m\times m}$. Without loss of generality, we can assume that entries of $A_{12}$ and $A_{21}$ are just zero or one. Thus, we need to find $[A_{12}]_{n\times k}$ and $[A_{21}]_{k\times n}$ such that $Supp(A_{12}A_{21})=Supp(A_1^*)$ and $k$ is minimum. We will show that the set basis problem \cite{Gary} can be reduced to the decision version of finding the minimal unobserved network which we call it the latent recovery problem. But before that, we define the set basis problem:

{\em The Set Basis Problem} \cite{Gary}: given a collection $\mathcal{C}$ of subsets of a finite set $U=\{1,\cdots,n\}$ and an integer $k$, decide whether or not there is a collection $\mathcal{B}\subseteq 2^U$ of at most $k$ sets such that for every set $C\in \mathcal{C}$, there exists a collection $\mathcal{B_C}\subseteq \mathcal{B}$ where $\bigcup_{B\in \mathcal{B}_C} B=C$.

Any instance of the basis problem can be reduced to an instance of latent recovery problem. To do so, we encode any set $C$ in collection $\mathcal{C}$ to a row of $A_1^*=A_{12}A_{21}$ where $i$-th entry is equal to one if $i\in C$, and otherwise zero. It is easy to verify that the rows of matrix $A_{21}$ correspond to sets in collection $\mathcal{B}$ if there exist a solution for the basis problem. Since the basis problem is NP-complete, we can conclude that finding the minimal unobserved network is NP-hard.

\section{Proof of Theorem \ref{NMth}}\label{th:NMth}
Consider a minimal unobserved network $G_{min}$. Pick any latent node $i^{\prime}$ which its in-degree or out-degree is greater than one. Let $V^-_{i^{\prime}}$ and $V^+_{i^{\prime}}$ be the sets of nodes that are going to and incoming from node $i^{\prime}$, respectively. We omit the node $i^{\prime}$ and create $|V_{i^{\prime}}^-|\times|V_{i^{\prime}}^+|$ latent nodes $\{i^{\prime}_{j^{\prime}k^{\prime}}| j^{\prime}\in V_{i^{\prime}}^-, k^{\prime}\in V_{i^{\prime}}^+\}$. We also add a direct link from node $j^{\prime}\in V_{i^{\prime}}^-$ to $i^{\prime}_{j^{\prime}k^{\prime}}$ and from $i^{\prime}_{j^{\prime}k^{\prime}}$ to $k^{\prime}\in V_{i^{\prime}}^+$ in order to be consistent with measurements. We continue this process until there is no latent node with in-degree or out-degree greater than one. Since there exists at most one path with length $k$ from any observed node to another observed node, the resulted graph is exactly equal to graph $G_0$. Hence we can construct the minimal graph $G_{min}$ just by reversing the process of generating latent nodes from $G_{min}$ to merging latent nodes from $G_0$. But the NM algorithm consider all the sequence of merging operations. Thus, $G_{min}$ would be in the set $\mathcal{G}_{out}$ and the proof is complete.

\end{document}